\documentclass[letterpaper]{article} %
\usepackage{aaai23}  %
\usepackage{times}  %
\usepackage{helvet}  %
\usepackage{courier}  %
\usepackage[hyphens]{url}  %
\usepackage{graphicx} %
\usepackage{natbib}  %
\usepackage{caption} %
\usepackage{algorithm}
\usepackage{algpseudocode}
\usepackage{amssymb}
\usepackage{amsmath}

\usepackage{xcolor}
\usepackage{booktabs}
\usepackage{multirow}
\definecolor{citecolor}{HTML}{0071bc}
\usepackage[colorlinks=true,linkcolor=red, citecolor=citecolor]{hyperref}  
\usepackage{newfloat}
\usepackage{listings}
\DeclareCaptionStyle{ruled}{labelfont=normalfont,labelsep=colon,strut=off} %
\floatstyle{ruled}
\newfloat{listing}{tb}{lst}{}
\floatname{listing}{Listing}

\title{FedFOR: Stateless Heterogeneous Federated Learning with\\First-Order Regularization}
\author {
    \equalcontrib{Junjiao Tian},
    \equalcontrib{James Seale Smith},
    Zsolt Kira
}
\affiliations {
    Georgia Institute of Technology\\
    \{jtian73,jamessealesmith,zkira\}@gatech.edu
}

\iftrue
    \newcommand
    {\revise}[1]{{#1}\normalfont}
\else
    \newcommand
    {\revise}[1]{{\color{blue}#1}\normalfont}
\fi

\begin{document}
\maketitle
\begin{abstract}
           Federated Learning (FL) seeks to distribute model training across local clients without collecting data in a centralized data-center, hence removing data-privacy concerns. A major challenge for FL is \textit{data heterogeneity} (where each client's data distribution can differ) as it can lead to weight divergence among local clients and slow global convergence. 
           The current SOTA FL methods designed for data heterogeneity typically impose regularization to limit the impact of non-IID data and are \textit{stateful} algorithms, i.e., they maintain local statistics over time. \revise{While effective, these approaches can only be used for a special case of FL involving only a small number of reliable clients. For the more typical applications of FL where the number of clients is large (e.g., edge-device and mobile applications), these methods cannot be applied, motivating the need for a \textit{stateless} approach to heterogeneous FL which can be used for any number of clients.} 
           We derive a first-order gradient regularization to penalize inconsistent local updates due to local data heterogeneity. Specifically, to mitigate weight divergence, we introduce a first-order approximation of the global data distribution into local objectives, which intuitively penalizes updates in the opposite direction of the global update. The end result is a \textit{stateless} FL algorithm that achieves 1) significantly faster convergence (i.e., fewer communication rounds) and 2) higher overall converged performance than SOTA methods under non-IID data distribution. \revise{\emph{Importantly, our approach does not impose unrealistic limits on the client size, enabling learning from a large number of clients as is typical in most FL applications}} Our code will be released at \url{https://github.com/GT-RIPL/FedFOR}.

\end{abstract}

\section{Introduction}

Deep learning models for machine learning applications are typically trained offline on a large, static dataset contained in a data-center. However, this is difficult for many applications which consume \emph{personal user data} on edge devices such as phones, tablets, and computers. Particularly, collecting this data to a central server is often \textbf{a privacy concern with serious ethical and legal issues}. Instead, federated learning (FL)~\cite{mcmahan2017communication} has emerged as a decentralized alternative to standard centralized practices. In FL, a global \emph{server} communicates with \emph{distributed clients} which have local access to the data of interest. In lieu of collecting private data in a centralized data-center, the FL server will \emph{aggregate model updates} while keeping all data at the client level.

While the concept of FL is highly inviting, in practice there are many serious challenges when learning from decentralized data. First, \emph{client data is non-IID!} Client data will be shifted from the true underlying data distribution, causing clients to learn interfering knowledge~\cite{sahu2018convergence,li2019convergence}. This shift can take place as a \emph{prior shift} (think: clients have different interests, such as the number of dog pictures on a mobile phone for a dog owner) or as a \emph{covariate shift} (think: clients generate data in different styles or formats, such as different camera qualities on a mobile phone). 
\begin{figure}[t!]
    \centering
    \includegraphics[width=0.47\textwidth]{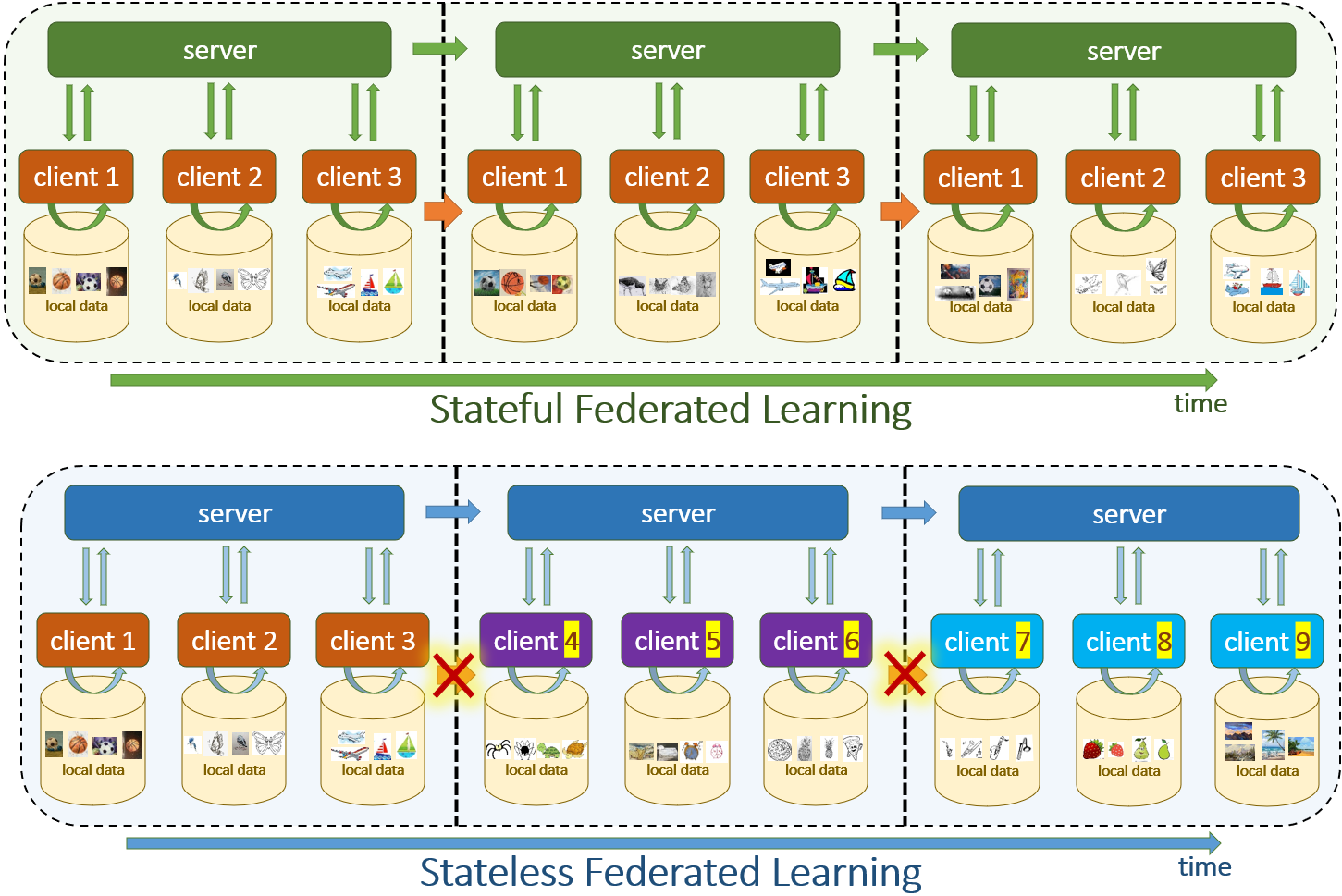}
    \caption{Heterogeneous federated learning involves distributed learning from clients which see different data distributions. Unlike prior works which propose \emph{stateful} algorithms which assume access to the same set of clients at every server update (top), our approach is \emph{stateless} and can learn from new clients at each server update (bottom). For large-scale federated learning with a large number of clients, the assumptions of the \emph{stateful algorithms fail}, thus motivating the \emph{need for stateless algorithms.}}
    \label{fig:key-idea}
\end{figure}
In our work, we therefore focus on the problem of non-IID client data, or \textbf{heterogeneous federated learning}.

The key challenge in such settings is that clients learning from heterogeneous data tend to \emph{diverge in their weight solutions}~\cite{zhao2018federated} given that the data distribution each client is learning from will invoke conflicting knowledge. Recent works mitigate this problem by maintaining local statistics at the client level to use for regularization schemes~\cite{zhang2020fedpd,acar2021federated}. 
\revise{As discussed in \citeauthor{kairouz2021advances}, these methods are characterized as \textit{stateful} algorithms (i.e., algorithms which maintain local statistics over time), and make a key assumption that clients can be regularly revisited, known as cross-silo FL. However, in \emph{practical, real-world heterogeneous FL applications} the number of clients is very large ($\sim10^{10}$) and thus \emph{clients infrequently participate} (i.e., train and send gradients back to the server), which is known as cross-device FL\footnote{Please see Appendix~\ref{app:statefullness} for a more detailed disucssion of statefullness in FL.}. Hence, these SOTA approaches for heterogenous FL cannot be scaled to most FL applications. It is instead desired to find a \textbf{stateless} approach to heterogeneous FL \emph{which can be used when the number of clients is large.}
}

Given this inspiration (see Figure~\ref{fig:key-idea}), we derive a first-order gradient regularization for \emph{stateless heterogeneous FL}. Intuitively, our method penalizes local updates that are inconsistent due to local data heterogeneity without leveraging persistent local statistics. At a technical level, our key insight is to include an approximated \textit{global objective} into local client objectives, using a first-order regularization derived from a Taylor  approximation of the global objective, hence needing only access to the global gradients as opposed to data from other clients or persistent local statistics. In addition to not requiring all local clients to maintain states over time, we even find that our method converges significantly faster than the current SOTA approaches across several heterogeneous FL benchmarks (i.e., our method converges with fewer communications between the server and the clients). Fast convergence is desired because it \emph{reduces the bandwidth} required for the FL system.

We also find that our method converges to the highest overall performance on the majority of benchmarks, compared to the main competing methods which make additional assumptions \emph{and} converge to lower overall performance. We particularly find that our method achieves the highest relative gains (both convergence \emph{and} performance) for experiments with a higher number of local training epochs. This makes sense because longer local training will lead to more severe interference between clients, increasing the impact of our global regularization scheme. Finally, we emphasize the importance of convergence in FL with a novel \emph{heterogeneous FL under concept shift} benchmark. The intuition for this experiment is that high convergence is not only important for reducing the bandwidth, but also important for fast recovery under concept shift. In summary, we make the following contributions:
\begin{enumerate}
    \item We contribute a stateless algorithm for heterogeneous FL which leverages global gradients from the server (always available) rather than local statistics from the clients (not available when the number of clients is large).
    
    \item We show that our approach achieves significantly faster convergence and higher overall accuracy than SOTA methods in both the prior shift setting and the covariate shift setting. The performance differential grows with the number of local training epochs, suggesting that our method is even more appealing under fewer global communication rounds.
    
    \item We highlight the importance of fast convergence for heterogeneous FL with a novel concept-shift benchmark, showing that the fast convergence of our method extends to high recovery under concept-shift.
\end{enumerate}

\section{Background and Related Work}

\noindent
\textbf{Federated Learning} (FL) was first introduced in FedAvg~\cite{mcmahan2017communication} to utilize unlimited user data without infringing on privacy. Broadly speaking, FL is a \textit{decentralized} machine learning setting, in which a global \textit{server} coordinates a large pool of local \textit{clients} to collectively solve a problem~\cite{zhang2021survey}. \textbf{Data heterogeneity} or non-IID data distributions is one of the biggest challenges in FL~\cite{zhang2021survey,kairouz2021advances}. This specifically refers to different data distributions on each client's local devices due to personal preferences and demographic differences. Distributed training on such diverged distributions can lead to weight divergence and poor aggregated performance, which becomes more severe as local training epochs increase. The vanilla FedAvg has been shown to suffer from data heterogeneity, both empirically~\cite{zhao2018federated} and theoretically~\cite{li2019convergence}.
\revise{To improve convergence under high data heterogeneity, exiting methods usually focus on two components in an FL algorithm: \textit{ServerOpt} and \textit{ClientOpt}~\cite{reddi2020adaptive}. They refer to optimizations performed locally on each client and optimization performed centrally on the server during aggregation, respectively. Most FL algorithms can be categorized under innovation to one of the components. \textbf{ServerOpt} methods improve convergence by adding server-side momentum or adaptive optimizers. FedAvgM~\cite{hsu2019measuring} and SlowMo~\cite{wang2019slowmo} simulates imbalanced data distribution, which is one form of data heterogeneity, and proposes to use server side momentum to improve convergence. FedAdagrad/FedYogi/FedAdam~\cite{reddi2020adaptive} extends FedAvg by including several adaptive optimizers on the server-side to combat data heterogeneity. \textbf{ClientOpt} methods add client-side regularization to reduce the effects of data heterogeneity. FedProx~\cite{li2020federated} proposes to add a proximal term (L2 regularization) to limit the impact of non-IID data. FedCurv~\cite{shoham2019overcoming} adapts a second-order gradient regularization method, EWC~\cite{kirkpatrick2017overcoming}, from continual learning to FL. Recent works, FedPD~\cite{zhang2020fedpd} and FedDyn~\cite{acar2021federated} improve convergence on non-IID data by including a first-order regularization term, which seeks consensus among clients. ClientOpt is investigated more heavily than ServerOpt because poor performance due to data heterogeneity is a direct result of local optimization. Furthermore, ClientOpt is more difficult to design because of the \textit{stateless} requirement for FL when the number of clients is large. For example, FedPD and FedDyn are \textit{stateful} algorithms and can be difficult to apply to large-scale, real-world FL settings. In this paper, we focus on the ClientOpt component to improve convergence under non-IID data distribution and compare analytically and empirically to all ClientOpt methods mentioned in this section. Federated learning is a fast growing field and there are many other works not discussed here due to space contraints. For example, FedBN~\cite{li2021fedbn} proposes to only partially sharing aggregated model and keep the local batch normalization layers unique to each client. This enables models to be more personalized while improving the overall aggregated performance. Its contribution is orthogonal to ours and we will show combined performance in our experiments. Another related FL technique is control-variate, e.g., SCAFFOLD~\cite{karimireddy2019scaffold}, which is also a \textit{statefull} algorithm. We give a detailed introduction to SCAFFOLD in Appendix~\ref{app:additional_related}.
}

\noindent

\section{Method}
\subsection{Federated Learning Setup}
\revise{
\label{sec:fl_intro}
Federated Learning (FL) is a distributed training paradigm~\cite{mcmahan2017communication}, where a global model (\textit{server}) is constructed by aggregating locally trained models on differing local data (\textit{clients}). In this section, we introduce basic concepts and notations in federated learning. Additionally, we frame the problem as an iterative optimization process, which will motivate our proposed method in Sec.~\ref{sec:proposal}. Specifically, on the $t$-th iteration, the global model $\mathbf{W}^{t-1}\in\mathbb{R}^d$ from the previous iteration is distributed to $K$ selected clients. Each client then optimizes an objective function (Eq.~\ref{eq:local_obj}) on its data $D_k$ for a predefined number of epochs $E$. 
\begin{align}
\label{eq:local_obj}
    \min_\mathbf{W} \mathcal{L}_k(\mathbf{W}) = \frac{1}{|D_k|} \sum_{\xi\in D_k} l(\mathbf{W};\xi)
\end{align}
Then, the updated local models $\mathbf{W}^t_k, \forall k$ are aggregated into a new global model, $\mathbf{W}^t$, minimizing the global objective in Eq.~\ref{eq:global_obj}, and the process repeats. We view FL as an \textit{iterative} process that optimizes on the local objectives and the global objective. In summary, the goal of FL is to minimize a global objective defined by a weighted sum of $K$ local objectives.
\begin{align}
\label{eq:global_obj}
    \min_\mathbf{W} \mathcal{L} = \frac{1}{K} \sum_{k=1}^K \mathcal{L}_k(\mathbf{W}),
\end{align}

where $\mathcal{L}_k(\mathbf{W})$ is the local objective on client $k$. Intuitively, during the local optimization step (Eq.~\ref{eq:local_obj}), the model on each client is trained on a different dataset $D_k$, potentailly sampled from very different local data distributions. This causes the local models to diverge in the weight space~\cite{li2019convergence} and results in poor performance when aggregated to minize the global objective (Eq.~\ref{eq:global_obj}).

\subsection{Proposed Method}
\begin{figure}[t!]
    \centering
    \includegraphics[width=0.45\textwidth]{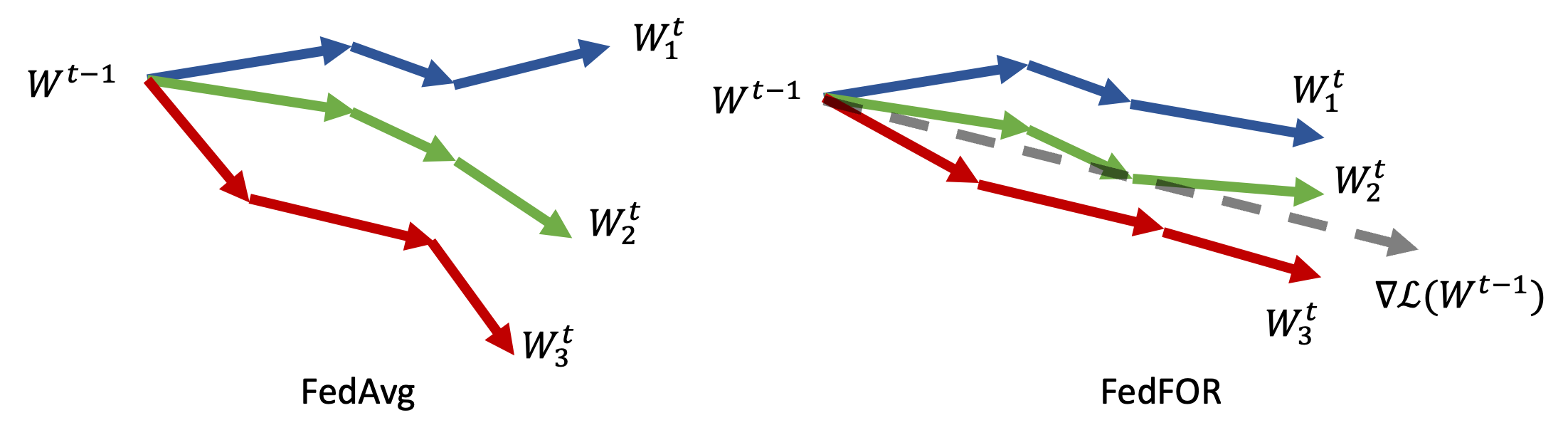}
    \caption{\textbf{Left}: Weight divergence due to data heterogeneity in FedAvg; \textbf{Right}: FedFOR induces a regularization to encourage local gradients to update in the same direction.}
    \label{fig:fedfor}
\end{figure}

\begin{algorithm}[t]
\caption{FedFOR. }\label{alg:alg1}
\begin{algorithmic}
\Require $K$ clients indexed by $k$ are selected; $B$ and $E$ represent local minibatch size and number of local training epochs; $\eta$ is the learning rate.

\State \textbf{Server executes:}
\State Initialize $\mathbf{W^0}$

\For{iteration $t=\{1,...T\}$ }
\For{$k\in\mathcal{K}$ in parallel}
\If{$t>1$}
\State $\mathbf{W}_k^t\leftarrow $ \textbf{ClientUpdate}$(\mathbf{W}^{t-1},\mathbf{W}^{t-2})$
\Else
\State $\mathbf{W}_k^t\leftarrow $ \textbf{ClientUpdate}$(\mathbf{W}^{t-1})$
\EndIf
\EndFor
\State $\mathbf{W}^t \leftarrow \frac{1}{K}\sum_{k=1}^K \mathbf{W}^t_k$
\EndFor\\

\State \textbf{ClientUpdate}$(\mathbf{W}^{t-1},\mathbf{W}^{t-2}= \text{None})$:\\
Create local batches $\mathcal{B}$ according to the batch size $B$
\For{$i= {1,...,E}$}
\For{$b\in\mathcal{B}$}
\If{$\mathbf{W}^{t-2}\neq \text{None}$}
    \State $\mathbf{W} \leftarrow \mathbf{W} -\eta \nabla \mathcal{L}_k^* $ (Eq.~\ref{eq:new_loss_3})
\Else
    \State $\mathbf{W} \leftarrow \mathbf{W} -\eta \nabla \mathcal{L}_k $ (Eq.~\ref{eq:local_obj})
\EndIf
\EndFor
\EndFor
\end{algorithmic}
\end{algorithm}

\label{sec:proposal}
 In this paper, we focus on the local optimization step, which minimizes local risk $\mathcal{L}_k$ (Eq.~\ref{eq:local_obj}). Due to the sequential nature, minimizing $\mathcal{L}_k$ without regard to the global data distribution or data from the other clients can lead to poor aggregated performance, especially when local data are non-IID. 

Even though access to data on other clients, which constitute the global objective, is prohibited in FL, \textbf{we propose to enhance the local objective with an approximated global objective without actual access.} Specifically, the enhanced local objective now optimizes two terms, 
\begin{align}
\label{eq:new_loss_1}
    \Tilde{\mathcal{L}_k} =  \mathcal{L}_k + \alpha\Tilde{\mathcal{L}} 
\end{align}

where $\Tilde{\mathcal{L}} $ is the approximated global objective and $\alpha>0$ is a hyperparameter.

To approximate the global objective, we use Taylor expansion to expand $\Tilde{\mathcal{L}}$ at $\mathbf{W}^{t-1}$, which is the previous global model, and only keep the first-order term:
\begin{align}
\label{eq:taylor}
   \Tilde{\mathcal{L}}(\mathbf{W}) = \mathcal{L}(\mathbf{W}^{t-1}) + \nabla_\mathbf{W} \mathcal{L}(\mathbf{W}^{t-1})^T (\mathbf{W}-\mathbf{W}^{t-1})\\\nonumber
    + \mathcal{O}(\mathbf{W}^2) + \mathcal{O}(\mathbf{W}^3) +... 
\end{align}

In Eq.~\ref{eq:taylor}, only the second term is related to the optimized variable $\mathbf{W}$. Thus, combining it with Eq.~\ref{eq:new_loss_1}, we can obtain the new objective without access to the global data:
 \begin{align}
 \label{eq:new_loss_2_temp}
      \Tilde{\mathcal{L}_k} =  \mathcal{L}_k + \alpha\nabla_\mathbf{W} \mathcal{L}(\mathbf{W}^{t-1})^T(\mathbf{W}-\mathbf{W}^{t-1})
 \end{align}
 
 Intuitively, by linearizing the global objective $\mathcal{L}(\mathbf{W})$ at a specific location $\mathbf{W}^{t-1}$, we effectively replace the need for access to global data with access to gradient at that location. Figuratively, as shown in Fig.~\ref{fig:fedfor}, FedFOR encourages local gradients that are consistent with the previous global update direction $\nabla_\mathbf{W} \mathcal{L}(\mathbf{W}^{t-1})$. The last step is to obtain the gradient vector $ \nabla_\mathbf{W}\mathcal{L}(\mathbf{W}^{t-1}) = \frac{1}{\eta}(\mathbf{W}^{t-1} - \mathbf{W}^{t})$, where $\eta$ is the learning rate. However, $\mathbf{W}^{t}$ is not available yet because it is the aggregated global model that has not been calculated. We use the gradient from the \textit{previous} global aggregation step to approximate it, \textit{assuming consecutive gradients are similar}\footnote{The assumption of similar consecutive updates stems from the common assumption of \textit{L-smooth} functions in FL~\cite{wang2021field}, where the loss function is assumed to be continuously differentiable, and its gradient is Lipschitz continuous with Lipschitz constant L, i.e. $\|\nabla f(x) - \nabla f(y)\|_2 \leq L\|x-y\|_2$.  If $x$ and $y$ are close, then their gradients are close. }. So the \textit{practical} version of Eq.~\ref{eq:new_loss_2_temp} follows,
 \begin{align}
   \label{eq:new_loss_2}
      \Tilde{\mathcal{L}_k} &=  \mathcal{L}_k + \alpha\nabla_\mathbf{W} \mathcal{L}(\mathbf{W}^{t-2})^T(\mathbf{W}-\mathbf{W}^{t-1})\\\nonumber
      & = \mathcal{L}_k + \frac{\alpha}{\eta}(\mathbf{W}^{t-2}-\mathbf{W}^{t-1})^T(\mathbf{W}-\mathbf{W}^{t-1})
 \end{align}
}
The linear term $\nabla_\mathbf{W}\mathcal{L}(\mathbf{W}^{t-2})^T(\mathbf{W}-\mathbf{W}^{t-1})$ can be viewed as an element-wise weighted regularization for the weights $\mathbf{W}$. Finally, we need to discuss the meaning of this regularization, especially its directionality. The regularization term is \textit{positive} when the previous gradient, $\nabla_\mathbf{W}\mathcal{L}(\mathbf{W}^{t-2})$, and the new \textit{local} update, $(\mathbf{W}-\mathbf{W}^{t-1})$ , are in the \textit{same} direction. In other words, the new local update is in the \textit{opposite} direction of the previous update because we subtract gradients when updating parameters. Minimizing this positive quantity on a client means penalizing \textit{opposing} local updates against previous global updates; when the regularization term is \textit{negative}, it means that the new local update is in the \textit{same} direction of the previous global update. Minimizing this negative quantity means encouraging updates in the same direction as previous updates, which can be outdated. We found that this encouragement leads to exploding negative gradient and does not contribute to better performance, so we only penalize opposing updates by only considering the positive components. We summarize the final loss function below and our federated learning algorithm, FedFOR, in Alg.~\ref{alg:alg1}. 

\begin{align}
\label{eq:new_loss_3}
    \mathcal{L}_k^*(\mathbf{W}) =  \mathcal{L}_k(\mathbf{W}) + \frac{\alpha}{\eta}\sum_{i=1}^{d} \mathcal{U}( (\mathbf{w}_i^{t-2}-\mathbf{w}_i^{t-1}) (\mathbf{w}_i-\mathbf{w}_i^{t-1}))
\end{align}
where $\mathcal{U}(x)=x,  \forall x\geq 0$ and $0$ otherwise. Note that the local update (Eq.~\ref{eq:new_loss_3}) does not require any local statistics. Therefore, FedFOR is a \textit{stateless} algorithm which does not need to maintain states on each client to keep track of global optimization iterations.

\revise{As like many prior works~\cite{gong2021ensemble,mcmahan2017communication,hsu2020federated,kairouz2021advances,li2021model,li2021fedbn}, this paper does not focus on convergence proof, because existing convergence analyses build on strong assumptions and apply only to the simplest FL settings such as convex functions. However, we can provide the following analysis and reference, which should give the audience confidence in the method's ability of faster convergence. Taking a derivative of the proposed FedFOR objective in Eq.~\ref{eq:new_loss_2} and applying it as in SGD, $$\textbf{W}^t_k =\textbf{W}^{t-1}_k - \eta\nabla\mathcal{L}_k(\textbf{W}) +\alpha(\mathbf{W}^{t-1}-\mathbf{W}^{t-2}),$$ gives us a distributed version of the widely used {Polyak momentum} update equation. A recent work~\cite{wang2021modular} shows that SGD with momentum has provably faster convergence rate than standard SGD in wide ReLU network and a deep linear  network. Therefore, we expect FedFOR with the same momentum update to have faster convergence, as shown in our experiments.}

\subsection{Comparisons to Prior Works}
Intuitively, our method (Eq.~\ref{eq:new_loss_3}) adds a first-order penalty to the vanilla loss function, which encourages new gradients to \textit{not} update in the opposite directions of the previous update. There are several notable works that share a similar high-level strategy, regularizing the weight updates, but differ in form. Specifically, we compare to FedAVG~\cite{mcmahan2017communication}, FedProx~\cite{zhao2018federated}, FedCurv~\cite{shoham2019overcoming} and FedPD~\cite{zhang2020fedpd}/FedDyn~\cite{acar2021federated}\footnote{We analyze FedPD and FedDyn together due to their similarity~\cite{zhang2021connection}.}. For example, the most related works, FedPD~\cite{zhang2020fedpd} and FedDyn~\cite{acar2021federated}, are \textit{stateful} algorithms whereas ours is \textit{stateless}. We will focus on how their local update strategy tackle non-IID data. FedAVG, as a baseline, adopts vanilla cross-entropy loss. 
\begin{align*}
    \mathcal{L}_{fedavg} = \mathcal{L}_k
\end{align*}
FedProx improves the vanilla formulation by adding a proximal term to limit the impact of local updates on differing distributions. As shown in Eq.~\ref{eq:fedprox}, this L2 regularization is \textit{uniformly} applied to all weights, whereas our method (Eq.~\ref{eq:new_loss_3}) is a \textit{weighted} regularization. 
\begin{align}
\label{eq:fedprox}
    \mathcal{L}_{fedprox} = \mathcal{L}_k + \frac{\alpha}{2} \|\mathbf{W} - \mathbf{W}^{t-1}\|^2_2
\end{align}

\revise{ FedCurv~\cite{shoham2019overcoming} proposes a second-order regularization technique inspired by the popular regularization method, EWC, in continual learning~\cite{kirkpatrick2017overcoming}. Specifically, the local objective of FedCurv takes the following form
\begin{align}
    \mathcal{L}_{fedcurv} = \mathcal{L}_k + \alpha\sum_{i\neq k} (\mathbf{W}-\mathbf{W}^{t-1}_i)^T\mathcal{I}_j^{t-1}(\mathbf{W}-\mathbf{W}^{t-1}_i).
\end{align}
where $\mathcal{I}_j^{t-1}$ is the diagonal fisher information matrix of client $j$ at the end of the previous round. However, EWC transfers poorly to federated learning because of the assumption of convergence. Specifically, EWC assumes that the previous round of local optimization (Eq.~\ref{eq:local_obj}) \textit{has converged}, which leads to $\nabla\mathcal{L}_j(\mathbf{W}_k^{t-1})=0$ and $\nabla^2\mathcal{L}_j(\mathbf{W}_k^{t-1})\approx\mathcal{I}_j^{t-1}$~\cite{martens2014new}. This assumption, that could be valid in the original continual learning setting, is no longer valid in FL because, due to limited computation, FL algorithms do not optimize the local model to convergence during every round of communication. Ignoring the validity of this assumption leads to sub-optimal performance of FedCurv as reported by a recent paper~\cite{xu2022acceleration}.
}

FedPD/FedDyn further improves the formulation by seeking consensus in local updates.  Specifically, FedPD derives its loss function from a global consensus reformulation of the original global objective (Eq.~\ref{eq:global_obj}).

\begin{align*}
\min_{\mathbf{W}_k,\mathbf{W}} \mathcal{L} = \frac{1}{K}\sum_{k=1}^K \mathcal{L}_k(\mathbf{W}_k) \quad & \text{s.t.} \quad \mathbf{W}_k =  \mathbf{W} \quad\forall k \in \mathcal{K}. 
\end{align*}

This reformulation enforces not only that local models, $\mathbf{W}_k$, optimize their own local objectives, but also that updated local models should converge to the \textit{same} global model $\mathbf{W}$. To solve this constrained optimization problem, augmented Lagrangian is used, which yields the following local loss function. 
\begin{align}
\label{eq:fedpd}
    \mathcal{L}_{fedpd} = \mathcal{L}_{k} + \nabla_{\mathbf{W}} \mathcal{L}_k(\mathbf{W}_k^{t-1})^T\mathbf{W} + \frac{\alpha}{2} \|\mathbf{W} - \mathbf{W}^{t-1}\|^2_2
\end{align}
\revise{FedPD/Dyn is most similar to our update strategy in Eq.~\ref{eq:new_loss_2} because both have a first-order gradient penalty term: $\nabla_{\mathbf{W}} \mathcal{L}_k(\mathbf{W}_k^{t-1})$ (Eq.~\ref{eq:fedpd}) is the gradient of the previous \textit{local} update; $\nabla_\mathbf{W}\mathcal{L}_k(\mathbf{W}^{t-2})$ (Eq.~\ref{eq:new_loss_2}) is the gradient of the previous \textit{global} update. The requirement of keeping track of local gradients from the previous iteration of communication makes FedPD/Dyn \textit{stateful} algorithms. For example, if the algorithm visits a different set of clients every time (cross-device FL), FedPD/Dyn degenerates to FedAvg because it does not have gradient information on these new clients and loses its speedup effects. In contrast, our algorithm is \textit{statelss} and applicable to a wider range of settings, in addition to faster convergence and better accuracy.}

\subsection{Communication Cost}
\label{sec:communication}
\begin{table}%
\caption{Comparisons of Server to Client (S2C) and Client to Server (C2S) communication cost. Note that under the cross-device FL setting, where each client is likely to be active only once, FedPD/DYN degenerates to FedProx}
\centering
\resizebox{1.0\linewidth}{!}{
    \begin{tabular}{c|c|c|c|c|c|c}  
        \toprule
        Method & Statefulness &\multicolumn{2}{c|}{Cross-Device FL} & \multicolumn{2}{c}{Cross-Silo FL} \\
        \midrule
        && S2C & C2S & S2C & C2S \\
        \midrule
        FedAVG& Stateless & $\mathbf{W}^{t-1}$ & $\mathbf{W}^{t}_k$  &$\mathbf{W}^{t-1}$ & $\mathbf{W}^{t}_k$ \\
       
        FedProx& Stateless & $\mathbf{W}^{t-1}$ &$\mathbf{W}^{t}_k$ & $\mathbf{W}^{t-1}$ & $\mathbf{W}^{t}_k$\\
       
        FedPD/DYN& Stateful & $\mathbf{W}^{t-1}$ &$\mathbf{W}^{t}_k$ &$\mathbf{W}^{t-1}$  & $\mathbf{W}^{t}_k$ \\
        
        FedFOR& Stateless& $\{\mathbf{W}^{t-1}$, $\mathbf{W}^{t-2}\}$ &$\mathbf{W}^{t}_k$ &  $\nabla_\mathbf{W}\mathcal{L}_k(\mathbf{W}^{t-2})$ & $\mathbf{W}^{t}_k$ \\ 
        \bottomrule
    \end{tabular}
}
\label{tab:communication}   
\end{table}

Communication is one of the most important design factors in FL research~\cite{mcmahan2017communication,kairouz2021advances}. We discuss it and compare our method to FedAvg~\cite{mcmahan2017communication}, FedProx~\cite{zhao2018federated} and FedPD~\cite{zhang2020fedpd}/FedDYN~\cite{acar2021federated} under each setting. We will discuss what information is transmitted from the server to clients (S2C) and what information is transmitted back from clients to the server (C2S) in Tab.~\ref{tab:communication}. For all algorithms, the previous global model $\mathbf{W}^{t-1} \in \mathbb{R}^d$ must be distributed to all selected clients, and all participating clients need to send back their update models $\mathbf{W}_k^t \in \mathbb{R}^d$. However, our algorithm, in the cross-device FL setting, when each client is only active in one round of communication, needs to send two consecutive previous global models to clients; under the cross-silo setting, we can reduce the communication cost by just sending the gradient from the previous step, $\nabla_\mathbf{W}\mathcal{L}(\mathbf{W}^{t-2})$, to local models. It is worth noting that, under the cross-device setting, FedPD/DYN (\textit{stateful}) degenerates to FedProx (\textit{stateless}) because the first order local gradient, $\nabla_{\mathbf{W}} \mathcal{L}_k(\mathbf{W}_k^{t-1})$, cannot be utilized until the second round of communication with the same clients. In summary, even though our method requires more communication bandwidth under the practical setting, as we will demonstrate later, it is a \textit{stateless} algorithm and has much stronger convergence behavior under distribution shift, and hence potentially requires less communication overall; under the cross-silo setting, our model has the same communication cost as others.

\section{Experiments}

\subsection{Datasets and Models}
\label{sec:datasets}
\textbf{Dataset.} To empirically demonstrate the efficiency of the proposed algorithm on non-IID data distributions, we use three benchmarks, consisting of both prior shift (Sec.~\ref{sec:prior_shift_exp}) and covariate shift (Sec.~\ref{sec:covariate_shift_exp}). For prior shift, we adopt the popular Imbalanced CIFAR10~\cite{cao2019learning} setting in imbalanced classification task.
For covariate shift benchmarks, following FedBN~\cite{li2021fedbn}, we use Digits, DomainNet~\cite{peng2019moment}, each of which consists of a range of different domains with shared labels. The Digits benchmark consists of SVHN~\cite{netzer2011reading}, USPS Hull~\cite{hull1994database}, SynthDigits~\cite{ganin2015unsupervised} and MNIST-M~\cite{ganin2015unsupervised}, MNIST~\cite{lecun1998gradient}; the DomainNet benchmark~\cite{peng2019moment} has six domains. We distribute the data from each domain to a client separately, such that each client has a distinct data distribution with covariate shift. Please see Appendix~\ref{app:implementation} for more details on implementation.

\textbf{Metrics.} For the majority of our experiments, we compare different methods using the accuracy achieved at both the halfway and full training iterations.
For the experiments on covariate shifts, we also analyze conference properties by reporting the global training steps required to achieve $X$ performance (given as ``Acc$x$"). Because each of the methods transmits the same number of models and the same model informative, this metric can be viewed as analyzing ``bandwidth" properties. For the experiments on concept shift (Sec.~\ref{sec:concept_shift_exp}), we reported average performance instead of performance at a few sampled places.

\subsection{Prior Shift Experiments}
\label{sec:prior_shift_exp}
\begin{table*}[h]
\caption{\textbf{Comparisons of Convergence on Prior Shifted Data with Imbalanced CIFAR-10 for \textit{Stateless} Algorithms.} The table reports the best validation accuracy up to the specified global iterations. Results for each method are reported in two rows, giving both mean (higher row) and standard deviation (lower row) over 3 runs. We present results for different number of local epochs $E\in\{1,2,4,8,16\}$. For each different $E$ setting, we report the halfway performance and the final performance using \% accuracy. Our algorithm, FedFOR, significantly outperforms competing methods, e.g., FedAvg, FedProx and FedCurv.}
\centering
\resizebox{1.0\linewidth}{!}{
 
    \begin{tabular}{l|c|c|c|c|c|c|c|c|c|c}  
        \toprule
        Local Epochs  &\multicolumn{2}{c|}{$E=1$} & \multicolumn{2}{c|}{$E=2$}  & \multicolumn{2}{c|}{$E=4$} & \multicolumn{2}{c|}{$E=8$} & \multicolumn{2}{c}{$E=16$}\\
        \midrule
        Global Iter. &  $t=100$ & $t=200$ & $t=100$ & $t=200$ & $t=50$ & $t=100$ & $t=50$ & $t=100$ & $t=25$ & $t=50$\\
        \midrule
        \multirow{2}{*}{FedAvg}  & 39.85 & 45.97 & 44.86	& 54.05 & 45.13 &	53.27 & 53.21 & 62.28 & 51.41 &	60.55\\
        & (1.02) & (0.79) & (1.38) & (0.58) & (0.01) & (0.61) & (0.95) & (0.52) & (0.23) & (0.96) \\
        \midrule
        
        \multirow{2}{*}{\textbf{FedProx }}
        & 39.84 &	46.04 &	44.82 &	53.98 &	44.68 &	52.62 &	51.88 &	62.17 &	52.90 &	60.66\\
        & (1.09) & (0.77) & (1.46) & (0.21) & (0.47) & (1.36) & (1.29) & (1.02) & (1.30) & (0.15) \\

        \midrule

        \multirow{2}{*}{\textbf{\textbf{FedCurv }}}
        &39.49&	46.03&	45.43&	53.36&	45.91&	53.68&	51.74&	60.94&	50.71&	60.97\\
        & (1.86) & (0.93) & (0.55) & (0.36) & (0.93) & (0.47) & (1.01) & (1.13) & (2.59) & (0.84) \\
        \midrule
      
        \multirow{2}{*}{\textbf{FedFOR }}
         &\textbf{41.93}&	\textbf{48.06}&	\textbf{50.18}&	\textbf{58.69}&	\textbf{52.18}&	\textbf{62.09}&	\textbf{60.24}&	\textbf{70.93}&	\textbf{56.86}&	\textbf{66.22}\\
        & (0.43) & (0.82) & (0.75) & (0.66) & (0.48) & (0.67) & (0.38) & (0.94) & (0.19) & (0.52) \\
        \bottomrule
    \end{tabular}
}
\label{tab:imbalance_cifar_ext}
\end{table*}

For prior shift experiments, we use Imbalanced CIFAR10~\cite{cao2019learning}. Specifically, we create a \textit{different} artificial long-tail distribution for each client's local data (Sec.~\ref{sec:datasets}). During each global iteration, we first randomly sample $10\%$ training data and further trim the data with a different long tail distribution using an imbalance ratio of $0.01$, i.e., each client possesses a different split of $\sim2.5\%$ of the training data. This simulates the real world FL \textit{stateless} scenario, where due to the large number of clients, each of the selected clients is likely to participate only \textbf{once} in a task~\cite{kairouz2021advances}. Notably, SOTA methods FedPD~\cite{zhang2020fedpd} and FedDYN~\cite{acar2021federated} are \textit{stateful} algorithms and therefore degenerate to FedAvg in this setting. Therefore, we compare to FedAvg~\cite{mcmahan2017communication}, FedProx~\cite{zhao2018federated} and FedCurv~\cite{shoham2019overcoming} in this section. 

Following conventional procedures~\cite{mcmahan2017communication,zhao2018federated}, we report results with different local training epochs, $E$. It's worth noting that there is a communication trade-off between the number of communication rounds (num. of global iterations) and the number of local update epochs~\cite{mcmahan2017communication}, i.e., the more local training epochs, the fewer global iterations are required. However, longer local training epochs lead to more severe weight divergence on non-IID data~\cite{zhao2018federated,li2021fedbn}. We show that our method not only consistently improves convergence, but also demonstrates larger performance differential with large $E$. In Tab.~\ref{tab:imbalance_cifar_ext}, we report validation accuracy results halfway through and at the end of training with $E\in\{1,2,4,8,16\}$. We observe that the proposed method converges both \textit{faster} and to a \textit{higher} accuracy. The largest performance differential is observed when $E=8$. Our method improves upon FedAvg relatively by $\sim15\%$ and $\sim16\%$ halfway through and at the end of training, respectively. We use the same hyperparameter $\alpha$ (Eq.~\ref{eq:new_loss_3}) in all experiments. Please see Appendix~\ref{app:hyperparameter} for details.

\subsection{Covariate Shift Experiments}
\label{sec:covariate_shift_exp}
\begin{table*}[h]
\caption{\textbf{Comparisons of Convergence on Covariant Shifted Data with 
DomainNet %
benchmark.} The table reports the best validation accuracy up to the specified global iterations. Results for each method are reported in two rows, giving both mean (higher row) and standard deviation (lower row) over 3 runs. We present results for different number of local epochs $E\in\{1,2,4,8,16\}$. For each different $E$ setting, we report the halfway performance and the final performance using \% accuracy. We also report the number of global iterations to reach 
$40\%$ %
under 
ACC40.} %
\centering

\resizebox{1.0\linewidth}{!}{
\begin{tabular}{c|c|c|c|c|c|c|c|c|c|c|c|c|c|c|c} 
\toprule
Local Epochs  &\multicolumn{3}{c|}{$E=1$} & \multicolumn{3}{c|}{$E=2$}  & \multicolumn{3}{c|}{$E=4$} & \multicolumn{3}{c|}{$E=8$} & \multicolumn{3}{c}{$E=16$}\\
\midrule
Global Iter. & ACC40 &  $t=100$ & $t=200$ & ACC40 &  $t=100$ & $t=200$ & ACC40 & $t=50$ & $t=100$ & ACC40 & $t=50$ & $t=100$ & ACC40 & $t=25$ & $t=50$\\ %
\midrule

\multirow{2}{*}{FedBN} & 46 & 45.87 &  \textbf{48.8} & 39 & \textbf{47.46} & 49.04 & 29 & 43.01 &  48.25 & 22 & 42.85 &  47.46 & 25 & 40.79 &  43.88 \\ 
 &  & (1.84) &  (1.75) &  & (1.42) &  (1.25) &  & (1.42) &  (1.58) &  & (1.51) &  (1.4) &  & (1.74) &  (1.78) \\ 
\multirow{2}{*}{FedProx} & 47 & 45.23 &  48.25 & \textbf{27} & 46.82 &  49.2 & 30 & \textbf{46.5} & 48.09 & 20 & 42.85 &  48.25 & 30 & 38.96 &  43.25 \\ 
 &  & (1.3) &  (1.47) &  & (1.58) &  (1.9) &  & (1.5) &  (1.28) &  & (1.75) &  (1.61) &  & (1.25) &  (1.9) \\ 
\multirow{2}{*}{FedDYN} & 57 & \textbf{46.5} & 48.41 & 38 & 46.5 &  48.57 & \textbf{21} & 45.87 &  \textbf{49.68} & 26 & 44.92 &  47.77 & 24 & 40.0 &  43.96 \\ 
 &  & (1.98) &  (1.81) &  & (1.9) &  (1.31) &  & (1.72) &  (1.89) &  & (1.68) &  (1.47) &  & (1.58) &  (1.11) \\ 
\multirow{2}{*}{FedCurv} & \textbf{44} & 44.76 &  48.09 & 32 & 47.3 &  48.73 & 29 & 42.69 &  45.71 & 22 & \textbf{45.39} & 49.04 & 33 & 33.96 &  44.92 \\ 
 &  & (1.1) &  (1.43) &  & (1.61) &  (1.91) &  & (1.96) &  (1.47) &  & (1.86) &  (1.26) &  & (1.8) &  (1.54) \\ 
\midrule 
\multirow{2}{*}{FedFOR} & 58 & 46.26 &  48.41 & 28 & 47.06 &  \textbf{49.76} & \textbf{21} & 44.76 &  48.09 & \textbf{17} & 44.36 &  \textbf{49.52} & \textbf{13} & \textbf{44.12} & \textbf{46.34} \\ 
 &  & (1.5) &  (1.01) &  & (1.71) &  (1.39) &  & (1.82) &  (1.66) &  & (1.03) &  (1.0) &  & (1.49) &  (1.86) \\

\bottomrule
\end{tabular}
}
\label{tab:covariate_domainnet_ext} %
\end{table*}
\begin{table*}[h]
\caption{\textbf{Comparisons of Convergence on Covariant Shifted Data with Digits benchmark.} The table reports the best validation accuracy up to the specified global iterations. Results for each method are reported in two rows, giving both mean (higher row) and standard deviation (lower row) over 3 runs. We present results for different number of local epochs $E\in\{1,2,4,8,16\}$. For each different $E$ setting, we report the halfway performance and the final performance using \% accuracy. We also report the number of global iterations to reach $80\%$ under ACC80.}
\centering

\resizebox{1.0\linewidth}{!}{
\begin{tabular}{c|c|c|c|c|c|c|c|c|c|c|c|c|c|c|c} 
\toprule
Local Epochs  &\multicolumn{3}{c|}{$E=1$} & \multicolumn{3}{c|}{$E=2$}  & \multicolumn{3}{c|}{$E=4$} & \multicolumn{3}{c|}{$E=8$} & \multicolumn{3}{c}{$E=16$}\\
\midrule
Global Iter. & ACC80 &  $t=50$ & $t=100$ & ACC80 &  $t=50$ & $t=100$ & ACC80 & $t=25$ & $t=50$ & ACC80 & $t=25$ & $t=50$ & ACC80 & $t=15$ & $t=25$\\
\midrule

\multirow{2}{*}{FedBN}  & 10 & 85.14 & 85.48 & 8 & 85.51 & 85.87 & 6 & 84.59 & 85.46 & 6 & 84.89 & 85.51 & 6 & 83.80 & 84.73\\ 
& & (0.23) & (0.05) & & (0.46) & (0.38) & & (0.16) & (0.13) & & (0.23) & (0.08) &  & (0.31) & (0.42) \\

\multirow{2}{*}{FedProx} & 10 &  85.04 & 85.59 & 7 & 85.43 & 85.95 & 6 & 84.89 & 85.67 &6& 84.60 & 85.75 & 7 & 83.42 & 84.52\\ 
&  & (0.13) & (0.20) &  & (0.17) & (0.18) &  & (0.53) & (0.17) & & (0.27) & (0.31) & & (0.06) & (0.24)\\

\multirow{2}{*}{FedDYN} &9 & 85.49 &	\textbf{85.85}	&7 & 85.26 &	\textbf{85.89}	&6 & 84.45 &	85.25	&5 &	84.88 &	85.63	&6 &	83.43 &	84.95\\
&  & (0.16) & (0.18) &  & (0.41) & (0.17) &  & (0.26) & (0.44) &  & (0.09) & (0.13) & & (0.80) & (0.30) \\

\multirow{2}{*}{FedCurv} & 10 & 85.16 & 85.56 & 7 & \textbf{85.79} & 86.10 & 7 & 84.34 & 85.53	 & 7 & 84.63 & 85.43 & 7 & 83.61 & 84.71\\
&  & (0.35) & (0.19) &  & (0.29) & (0.15) & & (0.01) & (0.12) &  & (0.41) & (0.25) &  & (0.19) & (0.48)\\

\midrule
\multirow{2}{*}{FedFOR} & \textbf{8} & \textbf{85.52} & 85.65 & \textbf{6} & 85.72 & 85.87 & \textbf{4} & \textbf{85.44} & \textbf{85.96 }& \textbf{4} & \textbf{85.72} & \textbf{86.12} & \textbf{4} & \textbf{84.94} & \textbf{85.58}\\
&  & (0.15) & (0.10) & & (0.17) & (0.27) & & (0.17) & (0.32) & & (0.35) & (0.06) & & (0.06) & (0.29) \\

\bottomrule
\end{tabular}
}

\label{tab:covariate_digits_ext}
\end{table*}

For covariate shift experiments, we report resutls for DomainNet in Tab.~\ref{tab:covariate_domainnet_ext} and Digits in Tab.~\ref{tab:covariate_digits_ext}. In our experiments, each domain is regarded as a separate client with different covariate shift. Instead of using FedAvg, we adopt the SOTA method FedBN~\cite{li2021fedbn} as the backbone for all compared methods. Specifically, we apply the local update rule from FedProx~\cite{zhao2018federated}, FedPD/DYN~\cite{zhang2020fedpd,acar2021federated}, FedCurv~\cite{shoham2019overcoming} and our method to FedBN. We also report results under various local update schedules. Again, we report halfway and final performance in the table. We observe that our model consistently outperforms competing methods with large local training epochs $E\in\{8,16\}$, when data heterogeneity causes more severe weight divergence~\cite{zhao2018federated}. This enables FedFOR to obtain higher communication efficiency because it can support more epochs of local updates and fewer rounds of global communication. As the number of local epochs increases, FedFOR demonstrates faster convergence compared to other methods. Most importantly, FedFOR remains \textit{stateless} while the closest competitor FedDyn/FedPD is \textit{stateful}.

\subsection{Concept Shift Experiments}
\label{sec:concept_shift_exp}
\begin{table}[t]
\caption{\textbf{Comparisons of Concept Shift Recovery on Covariant Shifted Data with DomainNet benchmark.} The table reports the average best validation accuracy over 200 training iterations with 10 randomly constructed concept shifts in the label space. We present results for different number of local epochs using \% accuracy}
\centering
\resizebox{1.0\linewidth}{!}{
\begin{tabular}{c|c|c|c|c|c} 
\toprule
Local Epochs  &\multicolumn{1}{c|}{$E=1$} & \multicolumn{1}{c|}{$E=2$}  & \multicolumn{1}{c|}{$E=4$} & \multicolumn{1}{c|}{$E=8$} & \multicolumn{1}{c}{$E=16$}\\
\midrule

FedBN & \textbf{44.51} & 46.84 & 48.01 & 47.55 & 47.48 \\ 
FedProx & 44.48 & 47.08 & 48.64 & 48.36 & 47.73 \\ 
FedDYN & 44.22 & 46.52 & 47.97 & 47.37 & 47.84 \\ 
\midrule
FedFOR & 44.36 & \textbf{47.28} & \textbf{49.35} & \textbf{49.12} & \textbf{49.83} \\

\bottomrule
\end{tabular}
}

\label{tab:concept-drift-domainnet_bn}
\end{table}
While existing works~\cite{zhang2020fedpd,zhao2018federated,li2020federated,li2021fedbn} have focused on prior shift and covariate shift, concept shift is largely ignored, which happens very frequently in the real world. For example, a word can take on different semantic meanings over time and its corresponding sentiment shifts. Fundamentally, our world is dynamic and FL algorithms, deployed on hundreds of thousands of devices, need to adapt quickly to a changing world. With this in mind, FL algorithms can really benefit from quick convergence when concept shift happens. A related setting, also called concept shift, has been explored by a prior work for client selection in FL~\cite{chen2020focus}. In their setting, concept shift is temporary and represents noise on local clients, and therefore, rejection method is studied. On the contrary, our interpretation of concept shift represents persistent and global shift to a new concept, and fast model convergence to the concept is required.  

To simulate the dynamic real world and our interpretation of concept shift, we implement a label shift strategy. Specifically, at the start of each global iteration when a server selects clients to participate in this round of communication, the label of a selected class of data, for all clients, will change to a different one in the set of available labels with a probability of $5\%$. Once changed, the label for this class will \textit{not} be reverted until being changed again in the future to a different label. This irreversible and global concept shift mimics the dynamism in the real world.  We apply this concept shift strategy to the DomainNet benchmark and compare our methods with SOTA methods with different local training schedules. In Tab.~\ref{tab:concept-drift-domainnet_bn}, we observe increasingly favorable results from the proposed method with increasing number of local training epochs. Specifically, FedFOR achieves the best performance for $E\in\{2,4,8,10\}$.

\section{Conclusions}

Data heterogeneity remains a major challenge in Federated Learning. While existing  best performing FL algorithms demonstrate promising results, they are stateful algorithms, which violates an important assumption in FL: each client is likely to participate only once or few times for a task. To alleviate the problem of weight divergence due to non-IID client data, we propose a principled \textit{stateless} federated learning algorithm, FedFOR. Specifically, we introduce a new local objective which includes an approximated global objective, and adopt a linearization strategy around the previous global model location to arrive at a first-order gradient regularization penalty. The new algorithm is tested on four non-IID FL benchmarks encompassing both prior and covariate shifts, and shows consistently faster convergence and better accuracy, even compared to stateful algorithms. Additionally, we propose a new realistic interpretation of concept shift in the real world, in which concepts evolve over time, and a new benchmark for testing robustness and recovery speed of FL algorithm in the face of sudden shifting to a new concept. Compared to other methods, FedFOR recovers and adapts to new concepts more quickly due to its faster convergence speed.

\bibliography{references}

\iftrue
    \clearpage
    \appendix
    \section*{Appendix}
    \section{Stateless vs. Stateful Algorithms}
\label{app:statefullness}
\textbf{Statefullness} is a characteristic of federated learning that depends on the scale of deployment. For \textit{cross-device} federated learning, where distribution scale can be up to $10^{10}$, the server is likely to only communicate with a client once (or at most a few times) because it would be rare to sample the same client given the size of the candidate pool~\cite{kairouz2021advances}. For \textit{cross-silo} federated learning, where the size of clients is $2$-$100$, each client can participate in multiple rounds of communications~\cite{kairouz2021advances}. The difference in distribution scale affects algorithm designs, i.e., whether clients can carry states from round to round. \textbf{States} are local statistics such as local gradient~\cite{karimireddy2019scaffold} and optimizer steps~\cite{reddi2020adaptive}, etc., which can affect subsequent optimization on the same device in the next round, if recorded. Therefore, cross-device FL algorithms need to be \textit{stateless} whereas cross-silo FL algorithms can be \textit{stateful}. For example, many algorithms highlighted in this paper are stateful algorithms~\cite{karimireddy2019scaffold,acar2021federated,zhang2020fedpd}, because they require passing local gradients from round to round on the same client. However, because the same client is rarely visited twice in cross-device FL, those algorithms largely degenerate to vanilla FedAvg~\cite{mcmahan2017communication} or FedProx~\cite{zhao2018federated} with a uniform L2 regularization. Therefore, considering \textit{statefullness} in algorithm design is critical to scaling up federated learning in real world setting~\cite{reddi2020adaptive,kairouz2021advances}, and our method FedFOR is designed to be stateless to scale up to potentially infinite number of clients (see the prior shift CIFAR experiments in 
Sec 4.2). In summary, being \textit{stateless} is a defining and important characteristic of large scale cross-device federated learning as emphasized by~\cite{kairouz2021advances,reddi2020adaptive}, failure to obey this requirement will significantly limit an algorithm's deployment in the real world.

\section{Additional Related Works}
\label{app:additional_related}
In the main paper, we discussed the local update schemes of several related works, FedAvg~\cite{mcmahan2017communication}, FedProx~\cite{zhao2018federated}, FedCurv~\cite{shoham2019overcoming} and FedDyn~\cite{acar2021federated}/FedPD~\cite{zhang2020fedpd} in details. Here, we introduce an additional FL algorithm that are closely related to ours because they introduce additional regularization for local update. 

SCAFFOLD~\cite{karimireddy2020scaffold} proposes to use local and global gradients to correct ``client-drift" in local update. Specifically, SCAFFOLD stores local gradients on each device and keeps track of global gradients, aggregated from all participating clients. 
\begin{align}
\label{eq:scaffold_1}
        \text{Local Gradient:} \quad c_k &= \nabla_{\mathbf{W}} \mathcal{L}_k(\mathbf{W_k}^{t-2})\\\nonumber
    \text{Global Gradient:} \quad c &= \nabla_\mathbf{W}\mathcal{L}(\mathbf{W}^{t-2})
\end{align}
Here, we only highlight the local update rule. The reader is referred to the original SCAFFOLD paper~\cite{karimireddy2020scaffold} for more details on how $c$ and $c_k$ are updated and maintained. The update is given as:

\begin{align}
\label{eq:scaffold_2}
    \mathbf{W}_k^{t} =  \mathbf{W}_k^{t-1} - \alpha \left(\nabla_{\mathbf{W}_k^{t-1}} \mathcal{L}_{fedavg} - c_k +c\right)
\end{align}
where $\alpha$ is a fixed learning rate. It is obvious from Eq.~\ref{eq:scaffold_1} and Eq.~\ref{eq:scaffold_2} that the algorithm is \textit{stateful} because it maintains local states $c_k$ from round to round.

\section{Effect of Performance vs $\alpha$.}
\label{app:hyperparameter}

\begin{figure}[h]
    \centering
    \includegraphics[width=0.32\textwidth]{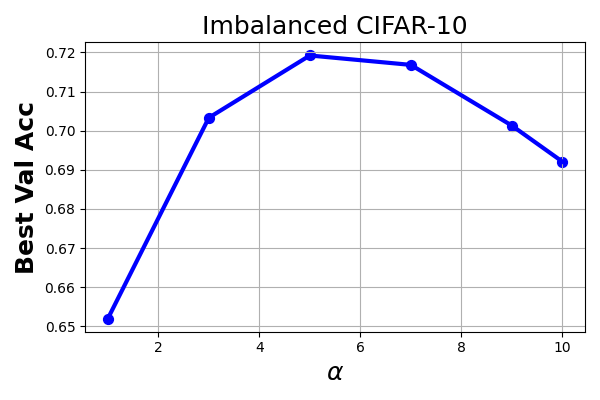}
    \caption{
    \textbf{Hyperparameter sweep of $\alpha$ for Imbalanced CIFAR-10 .} We use the same $\alpha=5$ for all experiments
    }
    \vspace{-2mm}
    \label{fig:hyperparameter-analysis}
\end{figure}
The choice of the hyperparameter affects performance of the proposed algorithm. Therefore, we conducted a parameter search for $\alpha$ using Imbalanced CIFAR for prior shift experiments with the number of local epochs $E=8$ as shown in Fig.~\ref{fig:hyperparameter-analysis}. We use the same $\alpha=5$ for all subsequent experiments.

\section{Implementation Details}
\label{app:implementation}
In this section, we will detail implementations of models and training details, organized by the order of experiments in the main paper: prior shift, covariate shifts and concept shift experiments. Following FedBN~\cite{li2021fedbn}, we use AlexNet for DomainNet, a custom six-layer CovNet~\cite{li2021fedbn} for Digits and ResNet20~\cite{he2016deep} for Imbalanced CIFAR10. 

In prior shift experiments, we use Imbalance CIFAR10~\cite{cao2019learning} and ResNet20~\cite{he2016deep}. Specifically, we use the proper ResNet implementation for CIFAR10~\cite{he2016deep}. The model is train with a batch size of 128 and SGD with a constant learning rate of 0.01 for varying number of global iterations depending on the number of local epochs, i.e, the more local update epochs the fewer global iterations. Note that to properly implement a \textit{stateless} algorithm, no SGD momentum and weight decay are used for all our experiments. In covariate shift experiments, we follow the setup from FedBN~\cite{li2021fedbn} using Digits, Office-Caltech~\cite{gong2012geodesic}, DomainNet~\cite{peng2019moment} benchmarks. We use AlexNet for DomainNet and Office-Caltech, a custom six-layer CovNet~\cite{li2021fedbn} for Digits. All models are trained with a batch seize of 32 and SGD with a constant learning rate of 0.01. In concept shift experiments, we use DomainNet and the custom six-layer ConvNet with the same optimization parameters as in the covariate shift experiments. Note that for all experiments, we use different combinations of global iterations and local update epochs, as indicated by the headers in each result table.

\fi

\end{document}